\documentclass[conference]{IEEEtran}
\IEEEoverridecommandlockouts

\usepackage{fix-cm}
\usepackage[UKenglish,USenglish]{babel}
\usepackage[OT1]{fontenc} 
\usepackage[utf8]{inputenc}

\usepackage[utf8]{luainputenc}

\usepackage[protrusion=true,expansion=true,tracking=true]{microtype}
\usepackage[style=USenglish]{csquotes}
\usepackage{textcomp}

\usepackage{cite}

\usepackage{amsmath,amssymb,amsfonts}
\usepackage{tocloft}
\usepackage{algorithmic}
\usepackage{float}

\usepackage{graphicx}
\graphicspath{{img/}}

\usepackage{array}
\usepackage{subcaption}
\usepackage{tabularx}
\usepackage{booktabs}
\usepackage{multirow}
\newcolumntype{Y}{>{\raggedleft\let\newline\\\arraybackslash\hspace{0pt}}X}
\newcolumntype{Z}{>{\centering\let\newline\\\arraybackslash\hspace{0pt}}X}

\usepackage{subcaption}
\usepackage{float}

\usepackage{url}
\usepackage{hyperref}
\usepackage{color}

\usepackage{enumitem}
\setlist[description]{noitemsep}
\setlist[enumerate]{noitemsep}
\setlist[itemize]{noitemsep,topsep=0pt}
\usepackage[acronym]{glossaries}
\def\BibTeX{{\rm B\kern-.05em{\sc i\kern-.025em b}\kern-.08em
    T\kern-.1667em\lower.7ex\hbox{E}\kern-.125emX}}

\usepackage[dvipsnames,svgnames,x11names]{xcolor}
\usepackage{colortbl}
\newcommand{\mycc}[1]{\cellcolor{gray!30}#1}

\begin{document}

\newcommand{\paligemma}{PaliGemma}
\newcommand{\llava}{LLaVA}
\newcommand{\qwen}{Qwen}
\newacronym{vllm}{VLLM}{Visual Large Language Model}
\newacronym{vllms}{VLLMs}{Visual Large Language Models}
\newacronym{llm}{LLM}{Large Language Model}
\newacronym{llms}{LLMs}{Large Language Models}
\newacronym{dpad}{PAD}{Presentation Attack Detection on ID Cards}
\newacronym{pad}{PAD}{Presentation Attack Detection}
\newacronym{idc}{IDC}{ID Card}
\newacronym{psp}{PSP}{Passport}

\title{Multimodal Models Meet Presentation Attack Detection on ID Documents}

\author{\IEEEauthorblockN{Marina Villanueva, Juan M. Espin}
\IEEEauthorblockA{\textit{Facephi} \\
Alicante, Spain, \\
marinavillanueva, jmespin{@facephi.com}}
\and
\IEEEauthorblockN{Juan E. Tapia}
\IEEEauthorblockA{\textit{da/sec-Biometrics and Security Research Group} \\
\textit{Hochschule Darmstadt, Germany}\\
juan.tapia-farias@h-da.de}
}

\maketitle

\begin{abstract}
The integration of multimodal models into Presentation Attack Detection (PAD) for ID Documents represents a significant advancement in biometric security. Traditional PAD systems rely solely on visual features, which often fail to detect sophisticated spoofing attacks. This study explores the combination of visual and textual modalities by utilizing pre\-trained multimodal models, such as \paligemma, \llava, and \qwen, to enhance the detection of presentation attacks on ID Documents. This approach merges deep visual embeddings with contextual metadata (e.g., document type, issuer, and date). However, experimental results indicate that these models struggle to accurately detect PAD on ID Documents.
\end{abstract}


\section{Introduction}
Multimodal models have gained prominence in recent years due to their ability to predict, classify, and describe images, audio, and text~\cite{Survey-FM}. These capabilities are supported by the large dataset used to train each network across multiple domains. These models, based on~\gls{llms}, may improve generalization across multiple domains that require access to restricted data. Most of these data are restricted to protect subjects' privacy, including biometric and demographic information, health data, and others~\cite{Survey-explainability}.

Today, several multimodal models are available in various versions, including~\paligemma \cite{steiner2024paligemma2}, \llava \cite{liu2023improvedllava},~\qwen~\cite{qwen}, and many others.

One of these challenge areas that needs to increase its generalization is \gls{pad} on ID Documents~\cite{IJCB2024-PAD, IJCB2025-PAD}, because it depends on new sets of bona fide ID Document images to improve its performance and extend to other countries. This restriction limited some companies' ability to expand their operations into new countries or continents. 

Gonzalez et al.~\cite{gonzalez2025forged} proposed a new serial multiclass, two-stage architecture that was trained from scratch without using any ImageNet weights. This architecture comprises one network that distinguishes bona fide ID cards from composite and synthetic cards, and a second network that distinguishes bona fide cards from those that are printed, displayed, or made of plastic. The dataset is private.

Motivated by the previous challenge, many companies are seeking to implement these multimodal approaches in their pipelines, based on successful reports and social media experience that demonstrate the models' ability to generate synthetic ID Documents and detect such attacks during remote onboarding~\footnote{\url{https://sl1nk.com/BFBbY}}. However, these capabilities have not been analyzed and reported in real datasets that include bona fide and attack images used in a commercial~\gls{pad} ID Document pipeline.

Our work benchmarks and inquiry three different~\gls{llms}:~\textit{PaLIGemma2\-3b\-mix\-224},~\textit{LLaVA1.6\-7b\-mistral}, and~\textit{Qwen2.5\-3b-instruct}. These three models were evaluated on their ability to predict whether an image is bona fide or an attack. For the purpose of this study, the analysis is restricted to ID cards and passports. We elaborated seven types of prompts to assess their generalization capabilities in real-world settings.

\textit{PaLIGemma2-3b-mix-224} is a compact vision-language model built on Google’s PaLI architecture. It is optimized for efficient multimodal understanding and operates at a resolution of $224\times224$ pixels, making it ideal for lightweight, real-time applications. 

\textit{LLaVA1.6-7b-mistral} is a larger model that has been instruction-tuned based on the Mistral-7B architecture. It has enhanced visual capabilities achieved through fine-tuning on image-text pairs, offering strong reasoning and instruction-following abilities in multimodal tasks. 

In contrast, \textit{Qwen2.5-3b-instruct} is a lightweight model from Alibaba that is also instruction-tuned. It is designed for high efficiency and strong language comprehension, focusing on providing clear, concise, and context-aware responses, even in complex or technical scenarios. 

While all three models support vision-language tasks,~\paligemma\ stands out for its compactness,~\llava\ is outstanding in reasoning, and~\qwen\ is particularly effective in instruction-following and fluency.

In order to evaluate their real contribution and assess their generalization capabilities, we evaluate the performance of these three models. The main contributions of this work are:

\begin{itemize}
\item It was to explore the application of three different~\gls{llms} for the task of~\gls{pad} on ID Documents, i.e., determine whether an ID Card/Passport is bona fide or an attack/spoof.

\item Seven different types of prompts organized as Single and Multiple prompts, prompts with examples, and prompts with background, and also task-oriented and recipe-style.

\item Our results indicate that the~\gls{llms} analyzed can not achieve generalization capabilities, which highlights the challenge to classify the~\gls{pad} on ID Documents correctly.
\end{itemize}

The article is structured as follows: Section~\ref{sec:art} reviews related work on ID card PAD systems. Section \ref{sec:metho} describes the methodology used based on multimodal models. Section~\ref{sec:metrics}~defines evaluation metrics. Section~\ref{sec:dataset}~describes the datasets. Section~\ref{sec:metho}~details the training methods and prompts. Section~\ref{sec:results}~presents experiments and results, and Section~\ref{sec:conclusions} concludes the work.

\section{Related Work}
\label{sec:art}

The use of~\gls{pad} on ID Documents has increased in recent years, driven by the widespread adoption of biometric remote onboarding systems, particularly as smartphones have become more prevalent globally. This system improves the user experience by eliminating the need to attend government or service offices physically. However, these services have increased the attempts to fake the system and impersonate users to gain economic benefit. The~\gls{pad} on ID Documents is an asymmetric challenge because the attacker requires only one pristine attack, whereas the defender requires millions of images to train the system.

Tapia et al.~\cite{IJCB2024-PAD, IJCB2025-PAD} have organized two state-of-the-art competitions to improve generalization capabilities. However, the results indicate that extending this~\gls{pad} on ID Cards system to multiple countries remains difficult due to the absence of publicly available data on genuine identity documents.

Previous attempts have analyzed the impact of LLMs across different image-based biometric modalities. Farmanifard et al.~\cite{ChatGPT-iris} analyze the capabilities of the~\textit{GPT-4} to explore its potential in iris recognition, which is a specialized field compared to the more common area of face recognition. By focusing on this challenge, they investigate how effectively AI tools such as ChatGPT can understand and analyze iris images.

Komatly et al.~\cite{ChatGPT-facepad} demonstrate the potential of ChatGPT (specifically~\textit{GPT-4o}) as a viable alternative for Face PAD, surpassing several PAD models, including commercial solutions based on few-shot learning.

Deandres-Tame et al.~\cite{ChatGPT-facebiom} evaluated the use of recent LLMs for the task of face verification, which involves determining whether two face images belong to the same individual. Specifically, they focused on the \textit{ChatGPT} chatbot and the latest multimodal~\textit{GPT-4}.

The \textit{PaLIGemma2-3b-mix-224},~\textit{LLaVA1.6-7b-mistral}, and~\textit{Qwen2.5-3b-instruct} are three distinct multimodal LLMs, each one designed with unique features and strengths.

\section{Methodology}
\label{sec:metho}
This section details the proposed methodology for evaluating different \gls{llms} for~\gls{pad} ID Document system. The strategy is structured in three stages: \textbf{1)} a study of the models' base knowledge on concepts related to ID Document fraud. \textbf{2)} the analysis of image descriptions, and \textbf{3)} concluding with the design of specific prompts meant to classify the authenticity of the ID Document presented. All the multimodal models have been tested locally hosted.
 
In the first stage, the models were asked about basic ID Document recognition terms to evaluate their theoretical grounding in attack detection, and the words most commonly used by each of the models to give definitions. The experiment involved the following prompts:

\begin{itemize}
    \item \textit{You're a forensic image analyst for a digital identity verification system. I want you to explain \textbf{what a presentation attack means} in the context of documental ID analysis}.
    \item \textit{You're a forensic image analyst for a digital identity verification system. I want you to explain what \textbf{types of presentation attacks} exist in the context of documental ID analysis and \textbf{classify them}.}
    \item \textit{You're a forensic image analyst for a digital identity verification system. I want you to explain \textbf{what a composite attack means} in the context of documental ID analysis.}
    \item \textit{You're a forensic image analyst for a digital identity verification system. I want you to explain what \textbf{types of composite attacks} exist in the context of documental {ID} analysis and \textbf{classify them}.}
\end{itemize}

In the second stage, the analysis of the models' knowledge base was deepened by requesting the description of ID Document images, both freely and with specifications related to~\gls{pad} on ID Documents field, to assess their visual understanding and implicit awareness of presentation attack cues. This qualitative assessment served as the foundation for the subsequent prompt engineering process. A \textbf{progressive prompting strategy} was employed to assess the depth of visual understanding:

\begin{enumerate}
    \item \textit{Describe the image.}
    \item \textit{You are an AI specialized in detecting fake or altered documental ID. Describe the image.}
    \item \textit{You are a forensic image analyst for a digital identity verification system. Describe the image.}
    \item \textit{Your task is to determine whether the document's ID is trying to deceive or not. Describe the image.}
    \item \textit{You are an AI specialized in detecting fake or altered document ID. The image provided is taken in the wild and might deceive a digital identity verification system. Describe the image, looking for signs of fraud, such as inconsistent shadows, edges, texture, semantic anomalies, text, or photo modifications.}
\end{enumerate}

Based on the observations derived from the description analysis, in the third stage, a diverse set of prompts was designed to progressively guide the models toward the PAD on ID Documents task. The prompts were grouped into several categories according to their structure and level of guidance as follows:

\begin{figure*}[]
    \centering
    \includegraphics[scale=0.3]{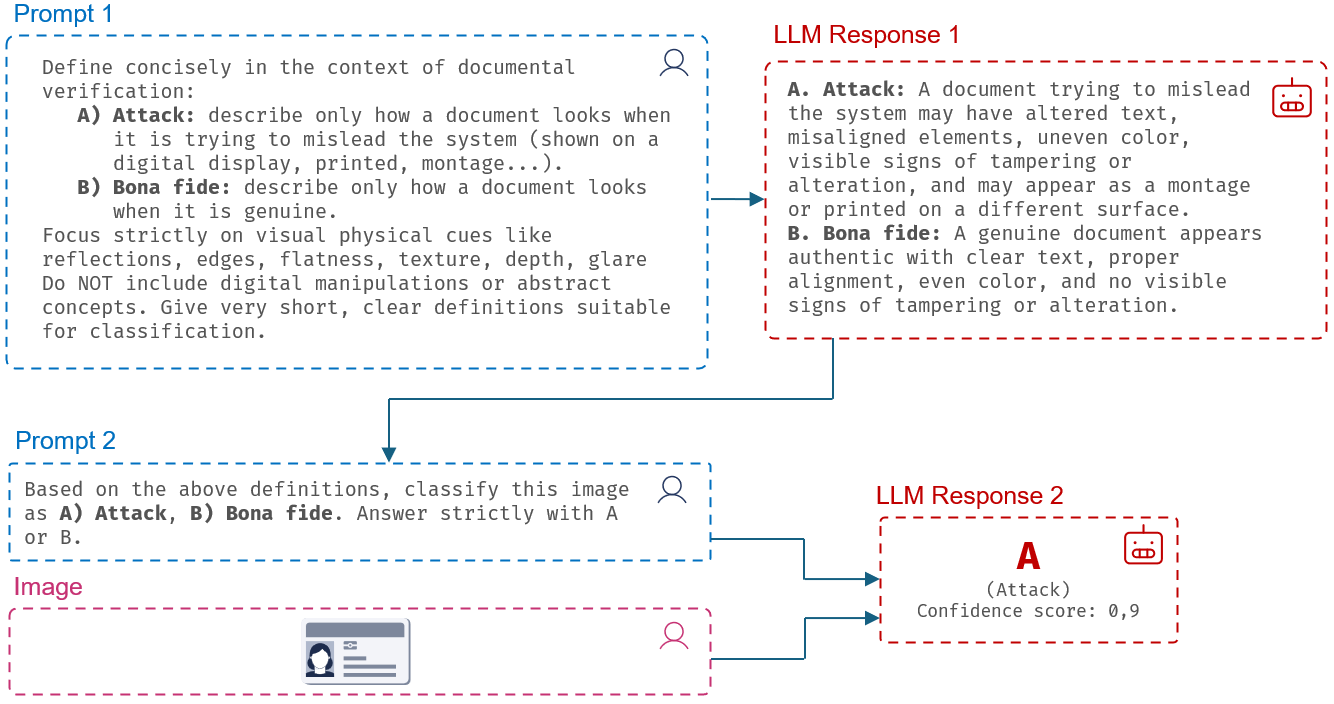}
    \caption{Multi-turn strategy.}
    \label{fig:multi_turn}
\end{figure*}

\begin{itemize}
    \item \textbf{Single-turn prompts}: the model is asked to classify the ID Document image as bona fide or spoof with the definitions given by a human expert. Example:~\enquote{\textit{You are a forensic image analyst for a digital identity verification system. Classify documental images into exactly one of the following categories: A) Attack: the document is trying to mislead the system (shown on a digital display, printed, montage...). B) Bona fide: a real document (natural depth, lighting, and background consistency). Answer strictly with one letter: A or B.}}

    \item \textbf{Multi-turn prompts}: the task is decomposed into sequential reasoning steps. First, the model is asked about specific definitions. Then, the model is asked to classify the ID Document image based on their previous definitions~(Figure \ref{fig:multi_turn}).

    \item \textbf{Simple prompts}: formulated as concise classification questions. Example: \enquote{\textit{Do you see a presentation attack in this image? Answer only with yes or no}}.
    
    \item \textbf{Prompts with examples}: providing reference cases to guide the model response (Example: \enquote{\textit{Does this image contain a 2D presentation attack (e.g., photo on a digital screen or a printed paper)? Answer with yes or no.}}
    
    \item \textbf{Prompts with background}, including contextual information about document fraud like real cases of spoofness, key parts of the images, hypothetical scenarios, and so on. Example:~\enquote{\textit{In the field of digital identity verification, there can be fraudsters who show up a fake or invalid document to steal information or money from another person. It is common to try to device these systems with images of documents shown on a screen (e.g., laptops, tablets, smartphones), printed on paper or PVC, with fields falsified...  It can be tricky to detect some spoofs, but make the biggest effort to try to detect the spoofness. Is this document likely to be rejected as spoofing? Answer strictly with yes or no.}}
    
    \item \textbf{Task-oriented prompts}, explicitly defining the role of the model as a PAD system. Example:~\enquote{\textit{You're a forensic image analyst for a digital identity verification system. Your work is to differentiate between bona fide and spoof document images. Is this image spoofing the system? Answer only with yes or no}}.
    
    \item \textbf{Recipe-style prompts}, describing the decision process step by step. Example:~\enquote{\textbf{Task}:~\textit{Document Presentation Attack Detection. Examine the provided documentary image carefully and perform a comprehensive fraud detection analysis.}\\ 
    \textbf{Step 1:} \textit{Systematic Visual Assessment  - Analyse the document layout, including alignment, spacing, and formatting consistency - Examine text quality (font consistency, printing quality, resolution)  - Check for visual artifacts (smudges, unusual erasures, pixelation, digital manipulation signs)  - Assess colour consistency, gradients and transitions across the document  - Identify signs of re-capture (Moiré patterns, screen glare, edge distortions, lighting reflections).} \\
    \textbf{Step 2:} \textit{Content Verification  - Examine personal information (name, address, dates) for completeness and consistency  - Check the validity and formatting of document issuance and expiry dates - Verify the presence, clarity, and expected appearance of official seals, holograms, or embedded security elements.} \\
    \textbf{Step 3:} \textit{Contextual Evaluation  - Check if document type, issuing authority, and security features are consistent with genuine specimens - Evaluate the quality and integration of holograms, microtext, or watermark patterns - Determine if the medium of presentation is consistent with authentic issuance (official polycarbonate/PVC vs. printed paper, digital screen display, or reprinted copy).}  \\
    \textbf{Step 4:} \textit{Potential Fraud Indicators  - Identify inconsistencies in numerical values - Detect evidence of photo substitution, tampering, or overlaid graphics - Detect misaligned text or irregular spacing patterns  - Detect if the document is displayed on an electronic screen (monitor, smartphone, tablet) - Detect if the document is printed on paper (standard, glossy, greyscale, color, etc.) - Detect if the document is a PVC/plastic reproduction - Detect if the image is a photograph or screenshot of another document - Note any unusual edits, erasures, or modifications   Conclusion: Based on your analysis, provide a decisive determination of the question: Is this identity document fraudulent? Answer with yes or no}}.
\end{itemize}

\begin{figure*}[]
    \centering
    \includegraphics[scale=0.29]{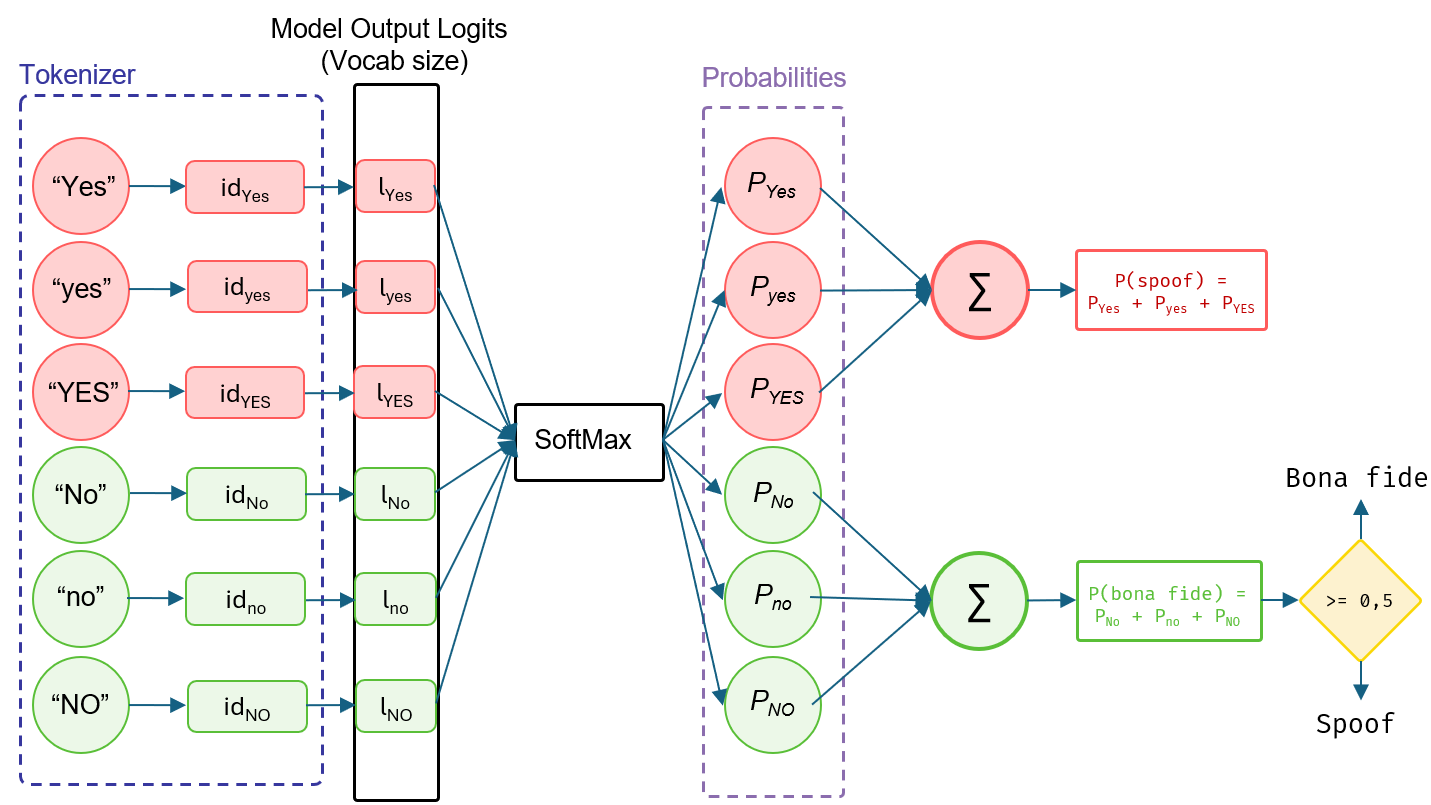}
    \caption{Scoring process based on model output logits applied to each model. Example for prompts with answer ``Yes'' or ``No''. }
    \label{fig:scoring}
\end{figure*}

\section{Metrics}
\label{sec:metrics}
This section defines the basic performance metrics for evaluating~\gls{pad} systems. Adherence to standardized metrics is essential to ensure that results are comparable and to drive progress in this field.

Figure \ref{fig:scoring} illustrates the method proposed to obtain the final response for each model. The process begins with the construction of the prompt, where the models are forced to respond following an ordered structure (e.g., “Yes”/“No” or “A”/“B”). This restricts the first token of the answer to a known token. After inference, the logits associated with these specific token IDs are extracted from the output layer. It is important to consider the handling of lexical variations of words. In this case, the semantic concept has been prioritized regardless of capitalization, so all morphological possibilities of the searched word are taken into account. (For example, \textit{“Yes”},~\textit{“yes”}, and~\textit{“YES”}). Then, a SoftMax function is applied to this subset of logits to normalize them and convert them into a valid probability function on the target tokens. 
Finally, the probabilities of tokens that share the same semantics are added together to obtain the final class score (e.g., $Score_{Genuine} = P_B + P_b$). The binary classification decision is determined by applying a standard threshold of $0.5$ to the genuine class score.

According to ISO/IEC 30107-3\cite{isoiec301073}, the PAD system has been evaluated according to different metrics. The Attack Presentation Classification Error Rate (APCER), which represents the proportion of attack presentations that have been incorrectly classified as bona fide presentations by a~\gls{pad} system at a specific operating threshold $\tau$ as is shown in Eq:\ref{eq:apcer}.

\begin{equation}\label{eq:apcer}
    {APCER_{PAIS}}=1 - \frac{1}{N_{PAIS}}\sum_{i=1}^{N_{PAIS}}RES_{i}
\end{equation}

The Bona fide Presentation Classification Error Rate (BPCER) represents the proportion of bona fide presentations that have been incorrectly classified as attack presentations by a \gls{pad} system at a specific operating threshold $\tau$, as is shown in Eq.\ref{eq:bpcer}.

\begin{equation}\label{eq:bpcer}
    BPCER=\frac{1}{N_{BF}}\sum_{i=1}^{N_{BF}}RES_{i}
\end{equation}

And the Equal Error Rate (EER), which is the trade-off between $APCER=BPCER$. Operational points such as BPCER\textsubscript{10} and BPCER\textsubscript{20} are usually reported for PAD in the state-of-the-art literature. For example, BPCER\textsubscript{10} corresponds to the BPCER when the APCER is 10\%, and BPCER\textsubscript{20} when the APCER is 5\%.

\section{Dataset}
\label{sec:dataset}

A \textit{bona fide image} is defined as a genuine image captured directly from the capture device without any modification, physically or digitally. An \textit{Attack image} is a print (ColorCopy), screen, PVC, or Physical Tamper version of a bona fide image presented at the moment of the onboarding process.

\begin{figure*}[]
    \centering
    \begin{subfigure}{0.37\textwidth}
        \centering
        \includegraphics[width=\linewidth]{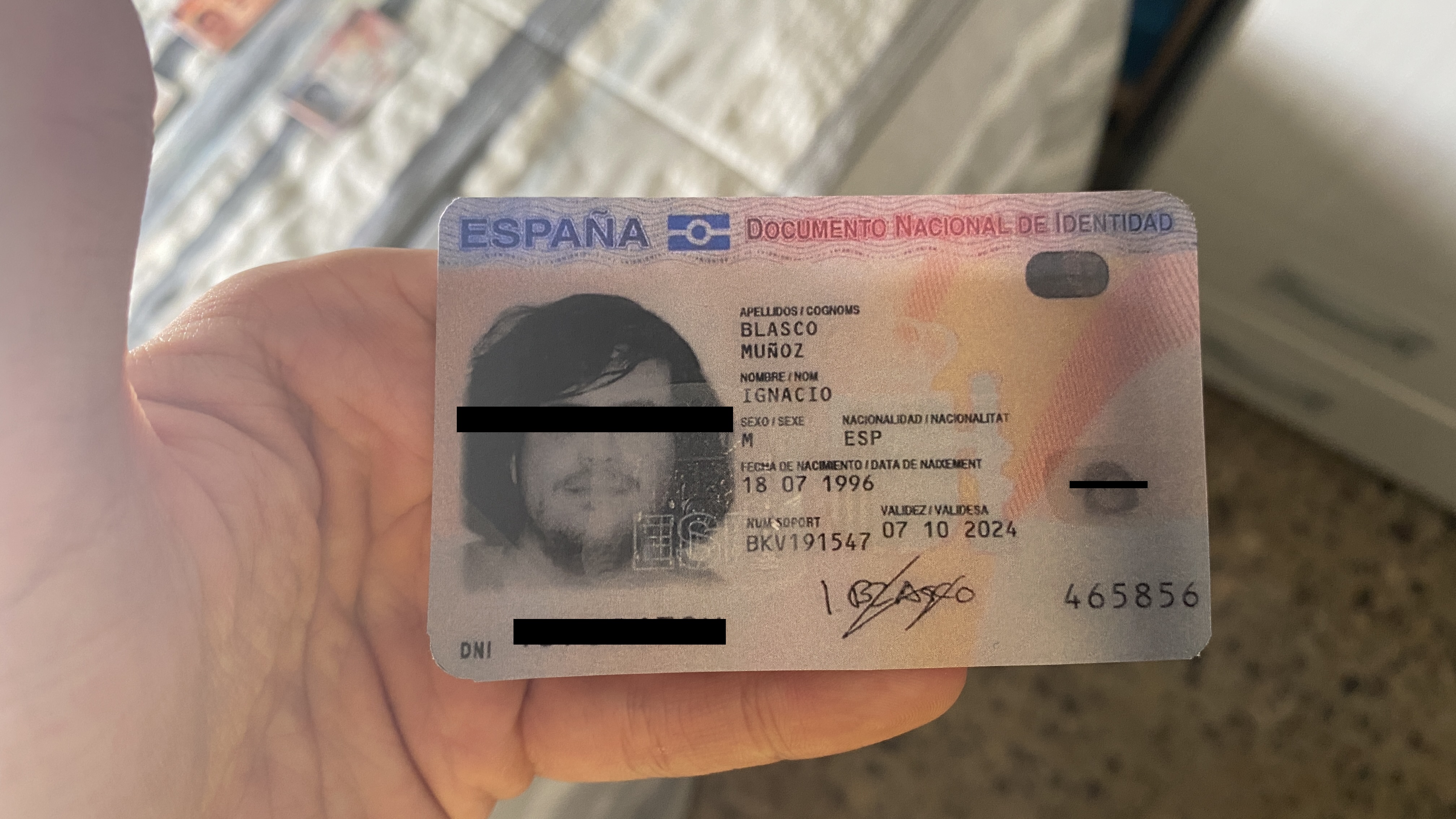}
        \caption{Paper attack}
        \label{fig:paper_attack}
    \end{subfigure}
    \begin{subfigure}{0.37\textwidth}
        \centering
        \includegraphics[width=\linewidth]{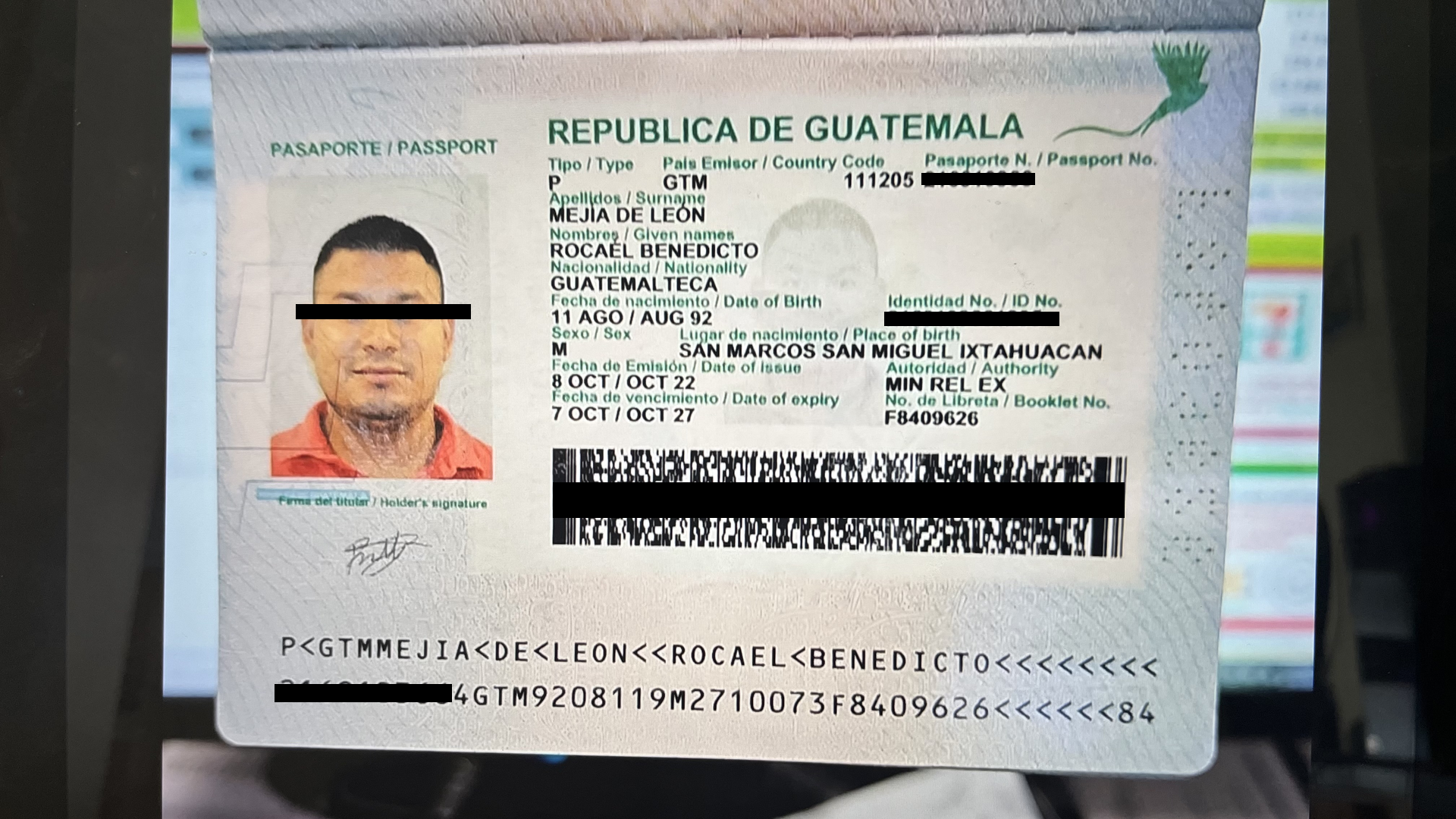}
        \caption{Screen attack}
        \label{fig:screen_attack}
    \end{subfigure} \\
    \begin{subfigure}{0.37\textwidth}
        \centering
        \includegraphics[width=\linewidth]{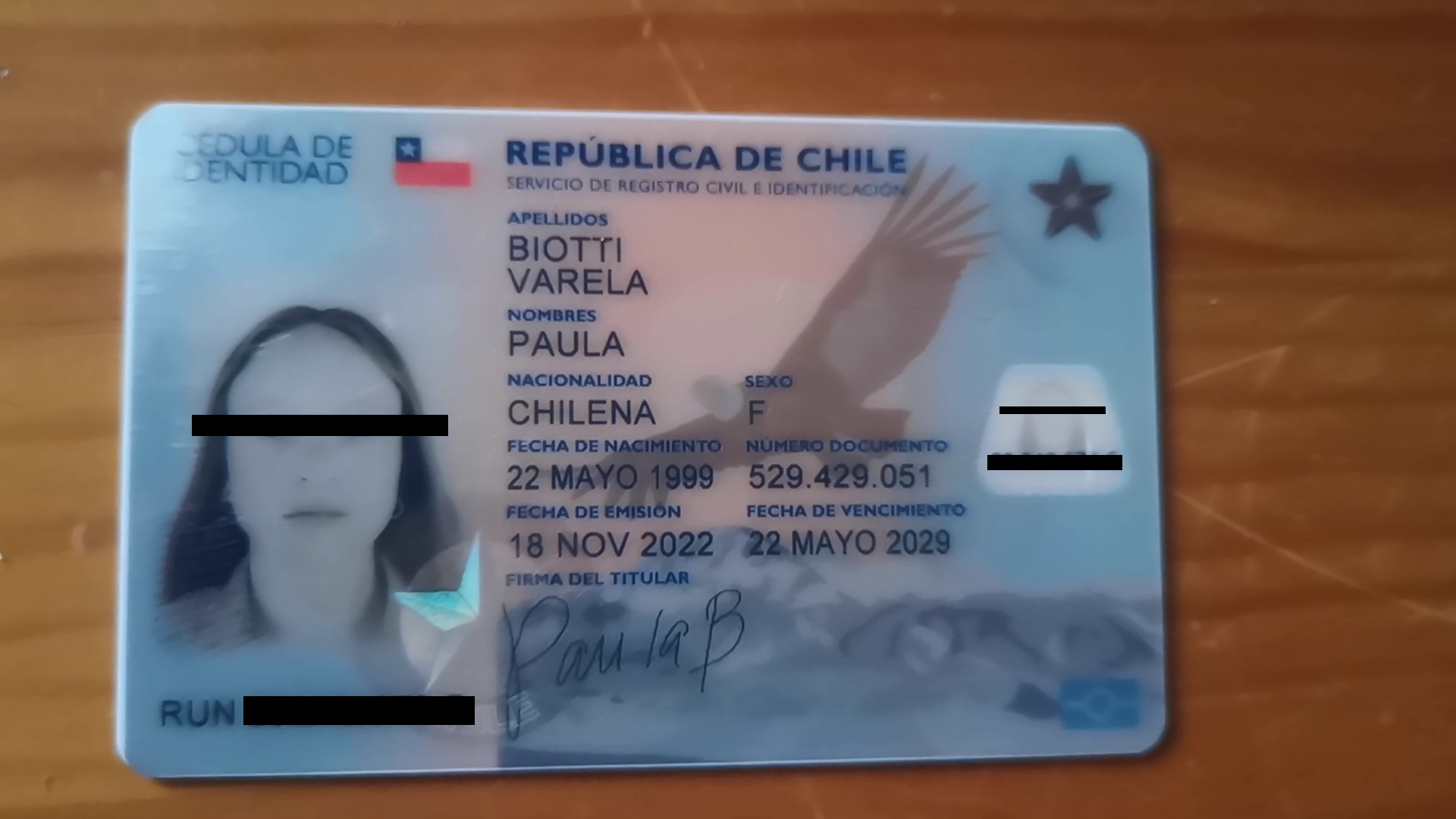}
        \caption{PVC attack}
        \label{fig:pvc_attack}
    \end{subfigure}
    \begin{subfigure}{0.37\textwidth}
        \centering
        \includegraphics[width=\linewidth]{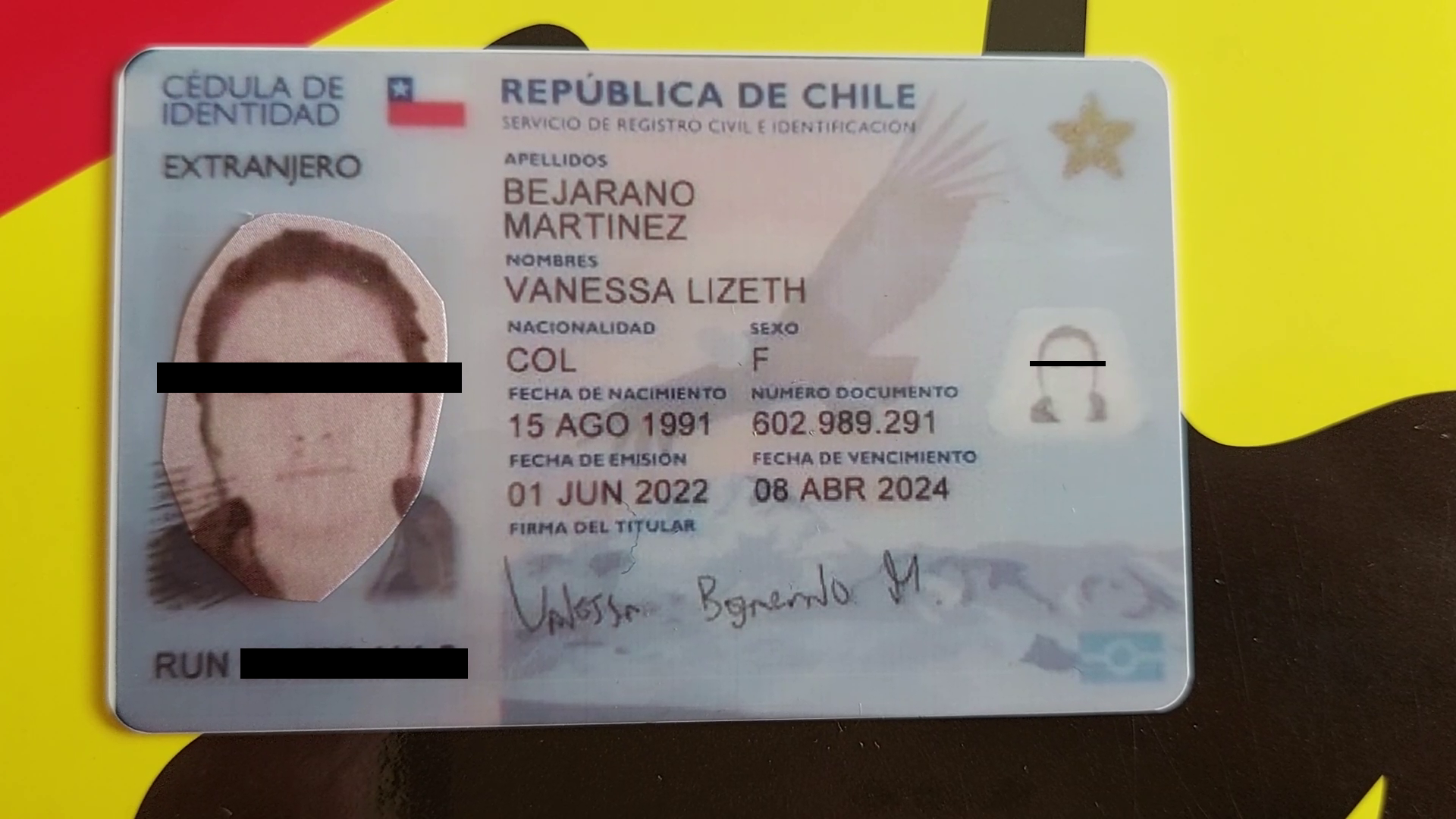}
        \caption{Tampered attack}
        \label{fig:tampered_attack}
    \end{subfigure}
    \caption{Attack examples for ID Cards and Passports.}
    \label{fig:attack_examples}
\end{figure*}

To address this challenge, two sets of $100$ images were used. A \textit{bona fide set} composed of $80$ \gls{idc} and 20 \gls{psp} from Guatemala (GTM), Chile (CHL), Spain (ESP), Nicaragua (NIC), Ecuador (ECU), El Salvador (SLV) and an \textit{attack set} composed of $85$ \gls{idc} and $15$ \gls{psp} from the same countries split between print ($25$), screen ($25$), PVC ($25$) and tampered ($25$) (Figure \ref{fig:attack_examples}). 

\section{Experiment and Results}
\label{sec:results} 

First, we examine the models' baseline knowledge related to document analysis and the most common types of attacks. Then, we analyze the models' inherent capabilities as a forensic image analyzer. Finally, we design seven different prompt styles and analyze the effectiveness of each one.

\subsection{Experiment 1. Definition Accuracy}\label{subsec:exp1}
In the first place, we analyzed the frequency of domain-specific keywords within the outputs generated by PaliGemma, LLaVA, and Qwen, performing an analysis similar to the one carried out in \cite{ChatGPT-fmad}. By counting the occurrences of these terms, we aimed to determine the semantic density of each model's response, measuring their capability to employ specialized vocabulary. The word list was categorized into 10 different classes, as follows:

\begin{enumerate}
    \item \textbf{Documents}:  \textit{passport}, \textit{card}, \textit{license}.
    \item \textbf{Identity}: \textit{ID}.
    \item \textbf{Bona fide}: \textit{genuine}, \textit{real}, \textit{authentic}, \textit{legit}.
    \item \textbf{Spoof}: \textit{deceive}, \textit{bypass}, \textit{fraud}, \textit{forgery}, \textit{attack}.
    \item \textbf{Manipulation}: \textit{tamper}, \textit{alter},  \textit{edit}, \textit{modify}.
     \item \textbf{Onboarding process}: \textit{verification}, \textit{authentication}, \textit{system}, \textit{security}.
    \item \textbf{Print attacks}: \textit{photo}, \textit{scan}, \textit{print}, \textit{copy}, \textit{physical}.
    \item \textbf{Screen attacks}:  \textit{computer}, \textit{pixel}, \textit{digital}, \textit{image}, \textit{video}.
    \item \textbf{Tamper attack}: \textit{composite}, \textit{layer}, \textit{splice}, \textit{combine}.
    \item \textbf{Artifacts}: \textit{inconsistency}, \textit{anomaly}, \textit{blur}, \textit{shade}.
\end{enumerate}

\begin{figure*}[]
    \centering
    \includegraphics[scale=0.35]{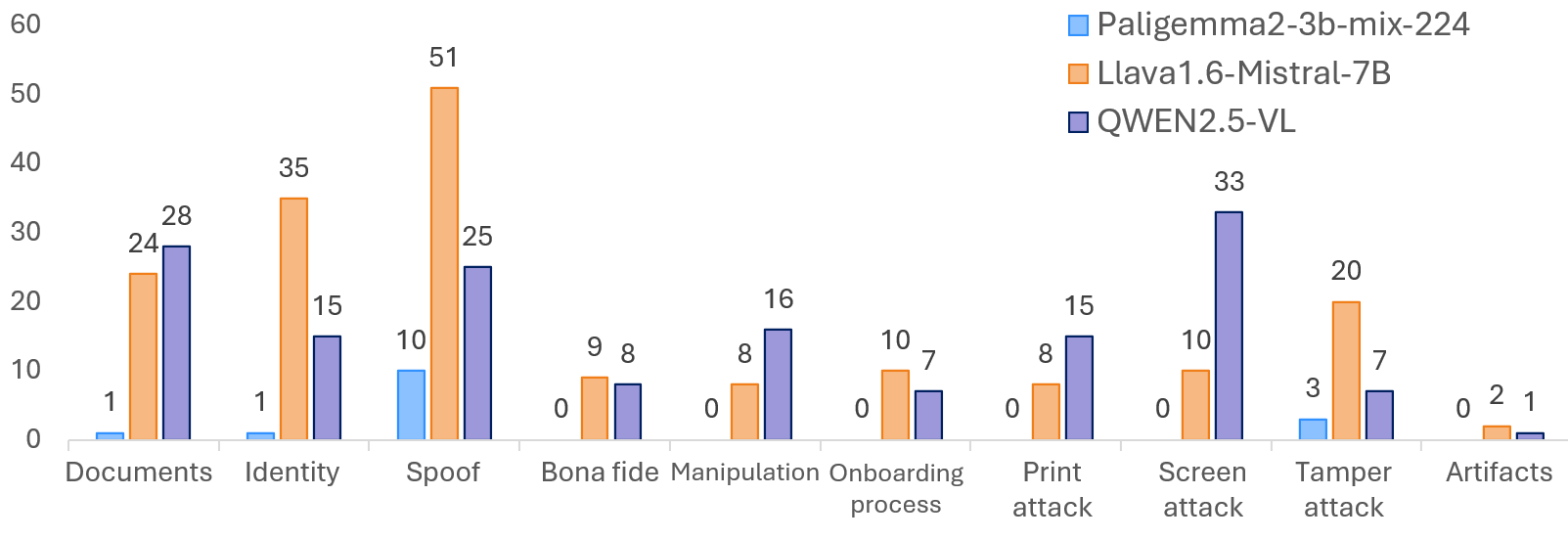}
    \caption{Results of Exp. 1. Comparison of the use of technical terminology among models.}
    \label{fig:frec}
\end{figure*}

The Figure \ref{fig:frec} illustrates the distribution of these keyword frequencies across the three LLM models. \paligemma\ demonstrated a lack of technical vocabulary and answers characterized by a high repeatability of generic terms or blank responses, rendering it ineffective for this task. In contrast, \llava\ and \qwen\ presented a stronger knowledge and understanding of domain terminology.

\subsection{Experiment 2. Visual Description}
\begin{figure*}[]
    \centering
    \includegraphics[scale=0.3]{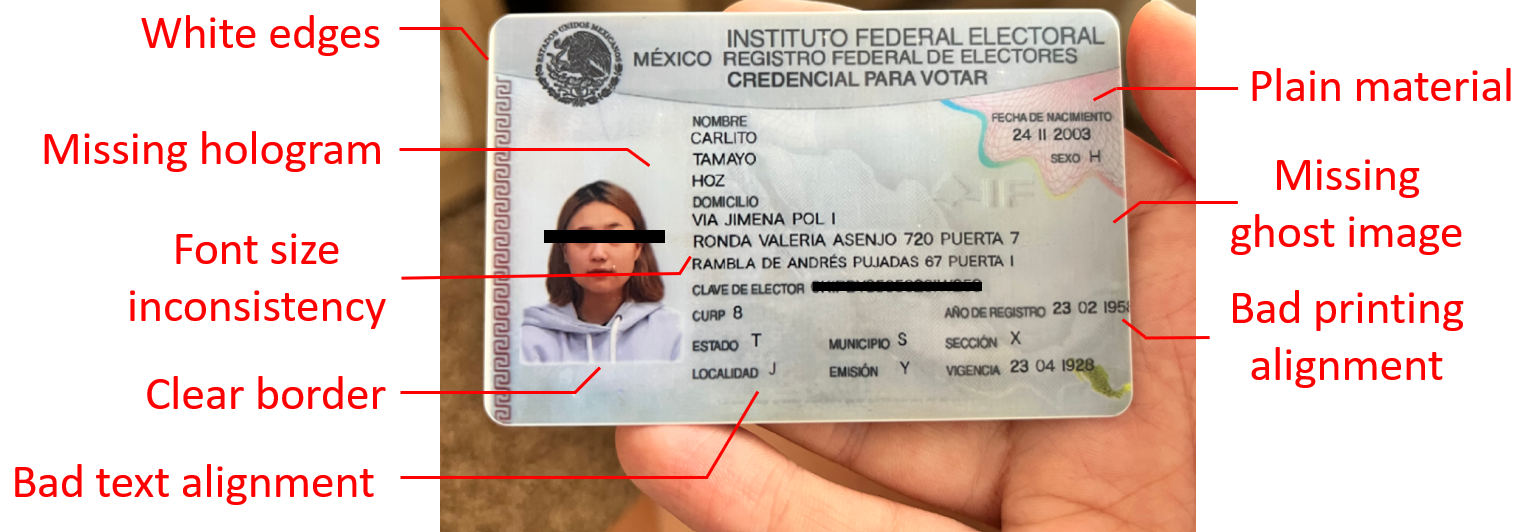}
    \caption{Visual keypoints of a spoof PVC ID Card.}
    \label{fig:pvc_keypoints}
\end{figure*}

This experiment aimed to evaluate their capability to apply this theoretical knowledge to a visual analysis task. To this end, the models were presented with a sample of a \textit{PVC attack}, a presentation attack instrument characterized by specific texture artifacts and specular reflections (Figure \ref{fig:pvc_keypoints}). 

Since in this experiment the prompts varied from a free description to a much more guided one, success was measured via a weighted scoring scheme depending on the degree of detection and autonomy of the models:
\begin{itemize}
    \item \textbf{3 points (Spontaneous Detection):} The model sees the flaw (e.g., “the texts are not aligned”) in the free phase without anyone telling anything. This demonstrates “forensic instinct”.
    \item \textbf{2 points (Guided Detection):} The model sees the flaw only when asked about a specific element.
    \item \textbf{1 point (Mention)}: The model talks about the element but does not take a position about its truthfulness.
    \item \textbf{0 points (Blindness):} The model does not see it even when asked.
    \item \textbf{\textit{H} (Hallucination):} The model sees it but gives an incorrect conclusion.
\end{itemize}

Table \ref{tab:descripciones} has many zero values, which means that current models do not see “physical defects”, they only read semantic content. Regarding response generation, both \qwen\ and \llava\ demonstrated the ability to produce increasingly elaborated descriptions. However, this increased verbosity in \qwen\ was accompanied by a higher incidence of hallucinations, whereas \llava\ showed a more stable performance.

\begin{table}[H]
\centering
\scriptsize
\caption{Results of Exp. 2. Visual Scores for PVC attack. Scores: 0 (Blindness), 1 (Mention), 2 (Guided detection), 3 (Spontaneous detection), H (Hallucination).}
\label{tab:descripciones}
\resizebox{\columnwidth}{!}{%
\begin{tabular}{lccccc}
\toprule
\textbf{Model} & \textbf{OCR} & \textbf{Security features} & \textbf{Material} & \textbf{Print quality} & \textbf{Total} \\ 
\midrule
\paligemma & 3 & 0 & 0 & 0 & \textbf{3}/12 \\
\llava     & 3 & 2 & 1 & H & \textbf{6}/12 \\
\qwen      & 3 & H & H & 0 & \textbf{3}/12 \\
\bottomrule
\end{tabular}
}
\end{table}

\subsection{Experiment 3. Prompt Design Strategy}
The last experiment ended with the development of 7 types of prompts that addressed the task of \gls{pad} on ID Documents from different angles. 

First, the performance of \gls{llms} as a classification system was studied based on the responses delivered. Table \ref{tab:summary_table} shows the APCER and BPCER values obtained. These results are equivalent to setting a threshold ($\tau$) of $0.5$ in Eqs. \ref{eq:apcer} and \ref{eq:bpcer}.


\begin{table*}[]
\centering
\scriptsize
\caption{Results of Exp. 3. N/A: Represents a null value because PaliGemma2 is not a multi-turn Chatbot. The results represents de APCER@0.5 and BPCER@0.5.}
\label{tab:summary_table}
\begin{tabular}{|c|cc|cc|cc|}
\hline
\multirow{2}{*}{Prompt} &
  \multicolumn{2}{c|}{Paligemma2 (\%)} &
  \multicolumn{2}{c|}{Llava1.6 (\%)} &
  \multicolumn{2}{c|}{QWEN2.5 (\%)} \\ \cline{2-7}
 &
  \multicolumn{1}{c|}{\textbf{APCER}} &
  \textbf{BPCER} &
  \multicolumn{1}{c|}{\textbf{APCER}} &
  \textbf{BPCER} &
  \multicolumn{1}{c|}{\textbf{APCER}} &
  \textbf{BPCER} \\ \hline
Single\_turn\_1 &
  \multicolumn{1}{c|}{100.0} &
  0.0 &
  \multicolumn{1}{c|}{0.0} &
  100.0 &
  \multicolumn{1}{c|}{98.0} &
  0.0 \\ \hline
Single\_turn\_2 &
  \multicolumn{1}{c|}{100.0} &
  0.0 &
  \multicolumn{1}{c|}{4.0} &
  73.0 &
  \multicolumn{1}{c|}{94.0} &
  4.0 \\ \hline
\mycc{Single\_turn\_3} &
  \multicolumn{1}{c|}{100.0} &
  0.0 &
  \multicolumn{1}{c|}{0.0} &
  100.0 &
  \multicolumn{1}{c|}{\mycc58.0} &
  \mycc{10.0} \\ \hline
Multi-turn 1 &
  \multicolumn{1}{c|}{N/A} &
  N/A &
  \multicolumn{1}{c|}{2.0} &
  98.0 &
  \multicolumn{1}{c|}{0.0} &
  100.0 \\ \hline
Multi-turn 2 &
  \multicolumn{1}{c|}{N/A} &
  N/A &
  \multicolumn{1}{c|}{0.0} &
  100.0 &
  \multicolumn{1}{c|}{0.0} &
  100.0 \\ \hline
Multi-turn 3 &
  \multicolumn{1}{c|}{N/A} &
  N/A &
  \multicolumn{1}{c|}{0.0} &
  100.0 &
  \multicolumn{1}{c|}{0.0} &
  100.0 \\ \hline
Simple\_1 &
  \multicolumn{1}{c|}{99.0} &
  0.0 &
  \multicolumn{1}{c|}{98.0} &
  1.0 &
  \multicolumn{1}{c|}{99.0} &
  0.0 \\ \hline
Simple\_2 &
  \multicolumn{1}{c|}{100.0} &
  1.0 &
  \multicolumn{1}{c|}{46.0} &
  37.0 &
  \multicolumn{1}{c|}{99.0} &
  0.0 \\ \hline
Simple\_3 &
  \multicolumn{1}{c|}{99.0} &
  0.0 &
  \multicolumn{1}{c|}{99.0} &
  0.0 &
  \multicolumn{1}{c|}{89.0} &
  7.0 \\ \hline
Simple\_4 &
  \multicolumn{1}{c|}{100.0} &
  0.0 &
  \multicolumn{1}{c|}{100.0} &
  0.0 &
  \multicolumn{1}{c|}{96.0} &
  2.0 \\ \hline
Simple\_5 &
  \multicolumn{1}{c|}{100.0} &
  0.0 &
  \multicolumn{1}{c|}{66.0} &
  29.0 &
  \multicolumn{1}{c|}{97.0} &
  0.0 \\ \hline
Simple\_6 &
  \multicolumn{1}{c|}{100.0} &
  0.0 &
  \multicolumn{1}{c|}{100.0} &
  0.0 &
  \multicolumn{1}{c|}{33.0} &
  51.0 \\ \hline
Simple\_7 &
  \multicolumn{1}{c|}{100.0} &
  0.0 &
  \multicolumn{1}{c|}{100.0} &
  0.0 &
  \multicolumn{1}{c|}{85.0} &
  16.0 \\ \hline
\mycc{Simple\_8} &
  \multicolumn{1}{c|}{100.0} &
  0.0 &
  \multicolumn{1}{c|}{88.0} &
  1.0 &
  \multicolumn{1}{c|}{\mycc73.0} &
  \mycc{17.0} \\ \hline
With\_examples\_1 &
  \multicolumn{1}{c|}{100.0} &
  0.0 &
  \multicolumn{1}{c|}{99.9} &
  0.0 &
  \multicolumn{1}{c|}{100.0} &
  0.0 \\ \hline
With\_examples\_2 &
  \multicolumn{1}{c|}{100.0} &
  0.0 &
  \multicolumn{1}{c|}{96.0} &
  6.0 &
  \multicolumn{1}{c|}{98.0} &
  0.0 \\ \hline
With\_examples\_3 &
  \multicolumn{1}{c|}{99.0} &
  0.0 &
  \multicolumn{1}{c|}{71.0} &
  0.0 &
  \multicolumn{1}{c|}{44.0} &
  31.0 \\ \hline
With\_examples\_4 &
  \multicolumn{1}{c|}{100.0} &
  0.0 &
  \multicolumn{1}{c|}{79.0} &
  10.0 &
  \multicolumn{1}{c|}{96.0} &
  1.0 \\ \hline
With\_examples\_5 &
  \multicolumn{1}{c|}{100.0} &
  0.0 &
  \multicolumn{1}{c|}{46.0} &
  17.0 &
  \multicolumn{1}{c|}{80.0} &
  6.0 \\ \hline
\mycc{With\_background\_1} &
  \multicolumn{1}{c|}{0.0} &
  100.0 &
  \multicolumn{1}{c|}{23.0} &
  60.0 &
  \multicolumn{1}{c|}{\mycc4.0} &
  \mycc{88.0} \\ \hline
With\_background\_2 &
  \multicolumn{1}{c|}{0.0} &
  97.0 &
  \multicolumn{1}{c|}{100.0} &
  0.0 &
  \multicolumn{1}{c|}{7.0} &
  89.0 \\ \hline
With\_background\_3 &
  \multicolumn{1}{c|}{0.0} &
  100.0 &
  \multicolumn{1}{c|}{100.0} &
  0.0 &
  \multicolumn{1}{c|}{100.0} &
  0.0 \\ \hline
With\_a\_task\_1 &
  \multicolumn{1}{c|}{100.0} &
  0.0 &
  \multicolumn{1}{c|}{100.0} &
  0.0 &
  \multicolumn{1}{c|}{36.0} &
  59.0 \\ \hline
With\_a\_task\_2 &
  \multicolumn{1}{c|}{100.0} &
  0.0 &
  \multicolumn{1}{c|}{58.0} &
  18.0 &
  \multicolumn{1}{c|}{2.0} &
  89.0 \\ \hline
As\_a\_recipe\_1 &
  \multicolumn{1}{c|}{100.0} &
  0.0 &
  \multicolumn{1}{c|}{0.0} &
  99.0 &
  \multicolumn{1}{c|}{0.0} &
  100.0 \\ \hline
\end{tabular}
\end{table*}

\paligemma\ exhibits a weak binary behavior. In most of the configurations evaluated, it fails to detect presentation attacks, systematically classifying inputs as bona fide samples. However, an opposite pattern was observed in \textit{With\_background} prompts. These prompts describe detailed fraud scenarios and provide explicit “tips” to identify potential manipulation. Under these conditions, the model reverses its decision pattern and systematically classifies samples as attacks. This inversion suggests that the model’s decisions are mainly driven by the semantic definition of the prompt rather than by a consistent visual analysis process.

\llava\ shows a slight attempt at analysis. Although it still dominates a binary behavior, the model achieves more balanced results in configurations such as \textit{With\_examples\_5} and \textit{With\_a\_task\_2}. However, performance still depends highly on the specific wording of the question.

Finally, \qwen\ presents the most reliable behavior among the evaluated models. The main difference from the previous ones is that it does not systematically fail at extreme APCER or BPCER values across all scenarios. In several \textit{Simple} and \textit{With\_examples} prompts, the model shows intermediate trade-offs between both metrics, suggesting the possible presence of discriminative reasoning and a stronger contribution of visual analysis in the classification process. However, in \textit{Multi\_turn} and \textit{As\_a\_recipe} prompts, \qwen\ displays a pattern similar to \llava\, which could indicate that the conversational structure may cause the models to be more likely to classify samples in a certain way.

Overall, the results obtained were not satisfactory, leading to doubts about whether the threshold of 0.5 had been a good calibration point. Therefore, secondly, the performance of \gls{llms} as regression systems was studied based on the method described in Figure \ref{fig:scoring}. This method serves to bypass output filtering mechanisms. By bypassing security protocols and post-processing constraints, this approach ensures consistent output availability across all three evaluated models.

Table~\ref{tab:summary_table_bpcer}, summarizes the results of the three multimodal models, based on EER, as well as two operational points BPCER\textsubscript{10} and BPCER\textsubscript{20}. These operational points are commonly used in commercial PAD systems and highlight the challenge associated with implementing and deploying multimodal models based on zero-shot and prompt-based approaches in onboarding systems. The best result for each model is highlighted in grey to compare the operational point with the binary answers in Table~\ref{tab:summary_table}. Although we obtained more realistic operational results, the EER, BPCER\textsubscript{10} and BPCER\textsubscript{20} showed low performance.

\begin{table*}[]
\centering
\scriptsize
\caption{Results of Exp. 3 evaluated at fixed BPCER operating points (BPCER\textsubscript{10} and BPCER\textsubscript{20}). N/A: Represents a null value because PaliGemma2 is not a multi-turn Chatbot. B10 and B20 represent the BPCER\textsubscript{10} and BPCER\textsubscript{20} respectively.}
\label{tab:summary_table_bpcer}
\begin{tabular}{|c|ccc|ccc|ccc|}
\hline
\multirow{2}{*}{Prompt} &
  \multicolumn{3}{c|}{Paligemma2 (\%)} &
  \multicolumn{3}{c|}{Llava1.6 (\%)} &
  \multicolumn{3}{c|}{QWEN2.5 (\%)} \\ \cline{2-10} 
 &
  \multicolumn{1}{c|}{\textbf{EER}} &
  \multicolumn{1}{c|}{\textbf{B10}} &
  \textbf{B20} &
  \multicolumn{1}{c|}{\textbf{EER}} &
  \multicolumn{1}{c|}{\textbf{B10}} &
  \textbf{B20} &
  \multicolumn{1}{c|}{\textbf{EER}} &
  \multicolumn{1}{c|}{\textbf{B10}} &
  \textbf{B20} \\ \hline
Single\_turn\_1 &
  \multicolumn{1}{c|}{66.0} &
  \multicolumn{1}{c|}{95.0} &
  98.0 &
  \multicolumn{1}{c|}{46.5} &
  \multicolumn{1}{c|}{79.0} &
  80.0 &
  \multicolumn{1}{c|}{42.0} &
  \multicolumn{1}{c|}{73.0} &
  82.0 \\ \hline
Single\_turn\_2 &
  \multicolumn{1}{c|}{44.0} &
  \multicolumn{1}{c|}{90.0} &
  98.0 &
  \multicolumn{1}{c|}{31.0} &
  \multicolumn{1}{c|}{53.0} &
  65.0 &
  \multicolumn{1}{c|}{40.5} &
  \multicolumn{1}{c|}{75.0} &
  86.0 \\ \hline
\textbf{\mycc{Single\_turn\_3}} &
  \multicolumn{1}{c|}{53.0} &
  \multicolumn{1}{c|}{93.0} &
  97.0 &
  \multicolumn{1}{c|}{35.5} &
  \multicolumn{1}{c|}{62.0} &
  74.0 &
  \multicolumn{1}{c|}{\textbf{\mycc35.0}} &
  \multicolumn{1}{c|}{\textbf{\mycc85.0}} &
  \textbf{\mycc96.0}\\ \hline
Multi turn\_1 &
  \multicolumn{1}{c|}{N/A} &
  \multicolumn{1}{c|}{N/A} &
  N/A &
  \multicolumn{1}{c|}{64.0} &
  \multicolumn{1}{c|}{93.0} &
  97.0 &
  \multicolumn{1}{c|}{49.5} &
  \multicolumn{1}{c|}{87.0} &
  90.0 \\ \hline
Multi turn\_2 &
  \multicolumn{1}{c|}{N/A} &
  \multicolumn{1}{c|}{N/A} &
  N/A &
  \multicolumn{1}{c|}{59.0} &
  \multicolumn{1}{c|}{99.0} &
  100.0 &
  \multicolumn{1}{c|}{43.0} &
  \multicolumn{1}{c|}{82.0} &
  87.0 \\ \hline
Multi turn\_3 &
  \multicolumn{1}{c|}{N/A} &
  \multicolumn{1}{c|}{N/A} &
  N/A &
  \multicolumn{1}{c|}{50.0} &
  \multicolumn{1}{c|}{99.0} &
  100.0 &
  \multicolumn{1}{c|}{46.0} &
  \multicolumn{1}{c|}{90.0} &
  93.0 \\ \hline
Simple\_1 &
  \multicolumn{1}{c|}{36.5} &
  \multicolumn{1}{c|}{62.0} &
  68.0 &
  \multicolumn{1}{c|}{50.0} &
  \multicolumn{1}{c|}{78.0} &
  89.0 &
  \multicolumn{1}{c|}{50.0} &
  \multicolumn{1}{c|}{86.0} &
  87.0 \\ \hline
Simple\_2 &
  \multicolumn{1}{c|}{53.0} &
  \multicolumn{1}{c|}{94.0} &
  96.0 &
  \multicolumn{1}{c|}{45.5} &
  \multicolumn{1}{c|}{79.0} &
  87.0 &
  \multicolumn{1}{c|}{54.5} &
  \multicolumn{1}{c|}{94.0} &
  96.0 \\ \hline
Simple\_3 &
  \multicolumn{1}{c|}{48.0} &
  \multicolumn{1}{c|}{75.0} &
  82.0 &
  \multicolumn{1}{c|}{52.5} &
  \multicolumn{1}{c|}{91.0} &
  96.0 &
  \multicolumn{1}{c|}{51.0} &
  \multicolumn{1}{c|}{92.0} &
  99.0 \\ \hline
Simple\_4 &
  \multicolumn{1}{c|}{39.5} &
  \multicolumn{1}{c|}{87.0} &
  95.0 &
  \multicolumn{1}{c|}{42.0} &
  \multicolumn{1}{c|}{81.0} &
  83.0 &
  \multicolumn{1}{c|}{53.0} &
  \multicolumn{1}{c|}{87.0} &
  92.0 \\ \hline
Simple\_5 &
  \multicolumn{1}{c|}{50.0} &
  \multicolumn{1}{c|}{86.0} &
  92.0 &
  \multicolumn{1}{c|}{42.0} &
  \multicolumn{1}{c|}{83.0} &
  93.0 &
  \multicolumn{1}{c|}{42.5} &
  \multicolumn{1}{c|}{84.0} &
  91.0 \\ \hline
Simple\_6 &
  \multicolumn{1}{c|}{41.0} &
  \multicolumn{1}{c|}{74.0} &
  79.0 &
  \multicolumn{1}{c|}{32.5} &
  \multicolumn{1}{c|}{65.0} &
  73.0 &
  \multicolumn{1}{c|}{44.0} &
  \multicolumn{1}{c|}{76.0} &
  88.0 \\ \hline
Simple\_7 &
  \multicolumn{1}{c|}{45.0} &
  \multicolumn{1}{c|}{73.0} &
  79.0 &
  \multicolumn{1}{c|}{40.5} &
  \multicolumn{1}{c|}{73.0} &
  86.0 &
  \multicolumn{1}{c|}{47.0} &
  \multicolumn{1}{c|}{74.0} &
  78.0 \\ \hline
\textbf{\mycc{Simple\_8}} &
  \multicolumn{1}{c|}{30.5} &
  \multicolumn{1}{c|}{50.0} &
  51.0 &
  \multicolumn{1}{c|}{\textbf{\mycc25.0}} &
  \multicolumn{1}{c|}{\textbf{\mycc50.0}} &
  \textbf{\mycc55.0} &
  \multicolumn{1}{c|}{46.0} &
  \multicolumn{1}{c|}{81.0} &
  88.0 \\ \hline
With\_examples\_1 &
  \multicolumn{1}{c|}{49.0} &
  \multicolumn{1}{c|}{80.0} &
  86.0 &
  \multicolumn{1}{c|}{43.5} &
  \multicolumn{1}{c|}{92.0} &
  94.0 &
  \multicolumn{1}{c|}{45.0} &
  \multicolumn{1}{c|}{73.0} &
  82.0 \\ \hline
With\_examples\_2 &
  \multicolumn{1}{c|}{45.0} &
  \multicolumn{1}{c|}{70.0} &
  78.0 &
  \multicolumn{1}{c|}{47.0} &
  \multicolumn{1}{c|}{96.0} &
  98.0 &
  \multicolumn{1}{c|}{44.0} &
  \multicolumn{1}{c|}{79.0} &
  90.0 \\ \hline
With\_examples\_3 &
  \multicolumn{1}{c|}{42.0} &
  \multicolumn{1}{c|}{84.0} &
  90.0 &
  \multicolumn{1}{c|}{35.5} &
  \multicolumn{1}{c|}{81.0} &
  94.0 &
  \multicolumn{1}{c|}{39.0} &
  \multicolumn{1}{c|}{96.0} &
  98.0 \\ \hline
With\_examples\_4 &
  \multicolumn{1}{c|}{59.0} &
  \multicolumn{1}{c|}{89.0} &
  93.0 &
  \multicolumn{1}{c|}{45.0} &
  \multicolumn{1}{c|}{77.0} &
  90.0 &
  \multicolumn{1}{c|}{52.0} &
  \multicolumn{1}{c|}{83.0} &
  85.0 \\ \hline
With\_examples\_5 &
  \multicolumn{1}{c|}{50.5} &
  \multicolumn{1}{c|}{80.0} &
  89.0 &
  \multicolumn{1}{c|}{36.0} &
  \multicolumn{1}{c|}{74.0} &
  78.0 &
  \multicolumn{1}{c|}{38.0} &
  \multicolumn{1}{c|}{70.0} &
  83.0 \\ \hline
With\_background\_1 &
  \multicolumn{1}{c|}{31.0} &
  \multicolumn{1}{c|}{53.0} &
  69.0 &
  \multicolumn{1}{c|}{49.0} &
  \multicolumn{1}{c|}{78.0} &
  84.0 &
  \multicolumn{1}{c|}{43.0} &
  \multicolumn{1}{c|}{71.0} &
  84.0 \\ \hline
With\_background\_2 &
  \multicolumn{1}{c|}{28.0} &
  \multicolumn{1}{c|}{47.0} &
  59.0 &
  \multicolumn{1}{c|}{55.0} &
  \multicolumn{1}{c|}{89.0} &
  89.0 &
  \multicolumn{1}{c|}{50.0} &
  \multicolumn{1}{c|}{86.0} &
  93.0 \\ \hline
\textbf{\mycc{With\_background\_3}} &
  \multicolumn{1}{c|}{\textbf{\mycc26.0}} &
  \multicolumn{1}{c|}{\textbf{\mycc73.0}} &
  \textbf{\mycc73.0} &
  \multicolumn{1}{c|}{30.5} &
  \multicolumn{1}{c|}{52.0} &
  74.0 &
  \multicolumn{1}{c|}{46.0} &
  \multicolumn{1}{c|}{92.0} &
  97.0 \\ \hline
With\_a\_task\_1 &
  \multicolumn{1}{c|}{40.0} &
  \multicolumn{1}{c|}{72.0} &
  81.0 &
  \multicolumn{1}{c|}{29.5} &
  \multicolumn{1}{c|}{64.0} &
  67.0 &
  \multicolumn{1}{c|}{47.0} &
  \multicolumn{1}{c|}{79.0} &
  84.0 \\ \hline
With\_a\_task\_2 &
  \multicolumn{1}{c|}{45.0} &
  \multicolumn{1}{c|}{67.0} &
  68.0 &
  \multicolumn{1}{c|}{40.0} &
  \multicolumn{1}{c|}{61.0} &
  72.0 &
  \multicolumn{1}{c|}{38.0} &
  \multicolumn{1}{c|}{80.0} &
  85.0 \\ \hline
As\_a\_recipe\_1 &
  \multicolumn{1}{c|}{38.0} &
  \multicolumn{1}{c|}{71.0} &
  87.0 &
  \multicolumn{1}{c|}{38.0} &
  \multicolumn{1}{c|}{73.0} &
  77.0 &
  \multicolumn{1}{c|}{50.0} &
  \multicolumn{1}{c|}{77.0} &
  86.0 \\ \hline
\end{tabular}%
\end{table*}

In this case, \paligemma\ continues to demonstrate non-operational behavior in most of the prompts evaluated, with several cases of EER greater than 50\%, suggesting that for certain prompts, its response logic is reversed. These results are accompanied by significantly high BPCER\textsubscript{10} and BPCER\textsubscript{20} ($\geq 90\%$), which means that the model does not achieve an acceptable compromise between safety and usability.

\llava\ is the model that achieves the best results with the \textit{Simple\_8} prompt. It achieved an EER of 25.0\% and a BPCER\textsubscript{10} of 50\%, which are high results for real-world biometrics but significantly better than its competitors.

\begin{figure*}[]
    \centering
    \begin{subfigure}{0.3\textwidth}
        \includegraphics[width=\textwidth]{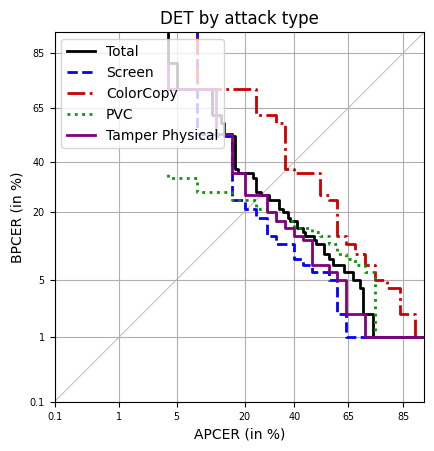}
        \caption{PaLIGemma2-3b-mix-224's-}
    \end{subfigure}
    \hfill
    \begin{subfigure}{0.3\textwidth}
        \includegraphics[width=\textwidth]{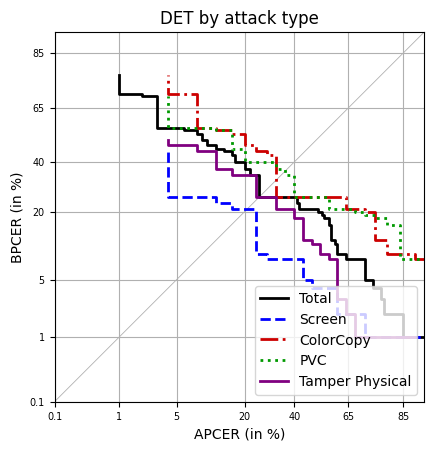}
        \caption{LLaVA1.6-7b-mistral's-}
    \end{subfigure} 
    \hfill
    \begin{subfigure}{0.3\textwidth} 
        \includegraphics[width=\textwidth]{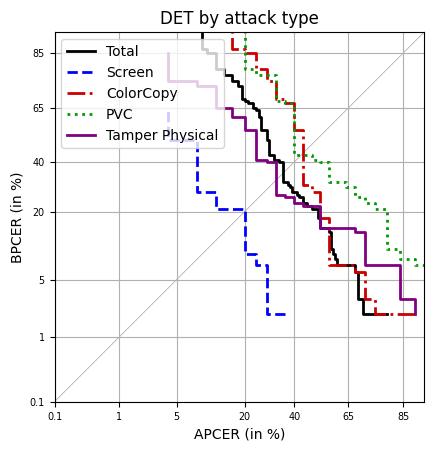}
        \caption{Qwen2.5-3b-instruct's.}
    \end{subfigure}
    \caption{DET performance achieved for the best prompts on the multimodal models (Left to Right): \textit{Single\_turn\_3}, \textit{Simple\_8}, and \textit{With\_background\_3}.}
    \label{fig:det}
\end{figure*}

Finally, \qwen\ is the model that presents the most consistent results across prompts; however, this consistency falls within an EER range of approximately 45\% to 55\%. This interval indicates performance close to random, reflecting limited discrimination capacity. Therefore, although it avoids extreme collapses such as those observed in \paligemma\, it also fails to reach operationally acceptable levels.

Figure \ref{fig:det} illustrates the best results based on the DET curve for a) PaLIGemma2-3b-mix-224, with the prompt \textit{\textbf{With\_background\_3}} (\enquote{\textit{A screen attack occurs when fraudsters present digital replicas of genuine documents on screens to deceive verification systems. The screen attack images are displayed on a variety of digital screens (e.g. laptops, tablets, smartphones) and photographed again in various resolutions. The print attacks are created using genuine documents printed out on standard paper, glossy paper and PVC cards with different thicknesses, resolutions, in grey scale or color. Composite attacks are generated by physically altering the user’s face in the PVC ID cards with paper cutouts of different faces, pasted in both rectangular and irregular shapes and photographing them again in various resolutions. Do you see any kind of attack in this image? Answer only with yes or no.}}), b) LLaVA1.6-7b-mistral with the prompt \textit{\textbf{Simple\_8}} (\enquote{\textit{Do you see a presentation attack in this image? Answer only with yes or no.}}), and c) Qwen2.5-3b-instruct models with the prompt \textit{\textbf{Single\_turn\_3}} (\enquote{\textit{You are a forensic image analyst for a digital identity verification system. Classify document images into exactly one of the following categories:}}).

\begin{enumerate}
    \item \textit{Screen Attack $\rightarrow$ the document is shown on a digital display (visible pixels, reflections, glare)}.
    \item \textit{Print Attack $\rightarrow$ the document is printed on paper or PVC (flat surface, edges, printing texture)}.
    \item \textit{Composite Attack $\rightarrow$ the document combines multiple images that are blended into different parts of the image (remove a signature, change the photo)}.
    \item \textit{Bona fide $\rightarrow$ a real document (natural depth, lighting, and background consistency)}.
\end{enumerate}
 
Screen attacks achieved the lowest EER among the three models. In contrast, the Printed-Color-Copy attack achieved the highest EER in all three models. 

\section{Conclusions}
\label{sec:conclusions}

This work presents an exhaustive assessment of three state-of-the-art multimodal models in a direct comparison for PAD on ID documents (ID cards and Passport). Our findings reveal a critical limitation in current multimodal models: despite their strong semantic understanding, they suffer from \enquote{physical blindness}—a failure to detect texture-based artifacts inherent in fake documents. This fundamental flaw undermines their reliability for real-world deployment. Furthermore, we identified a systematic bias across all the multimodal models, consistently classifying documents as bona fide, even when they are clearly manipulated. These results indicate that multimodal models are not yet ready to be entrusted with the critical task of detecting fake ID documents. 

As a future direction, we propose to investigate the relationship between model tokens and binarization results (e.g., A/B or Yes/No) in multimodal models. Our observations suggest that the model may interpret images differently based on tokenization—specifically, capitalization or font style. For instance, the model sometimes outputs answers in capital letters, which may correspond to distinct tokens in its vocabulary. While this distinction is semantically coherent for humans (e.g., \enquote{YES} vs. \enquote{yes}), it may lead to different token representations in the model’s internal processing. This could result in inconsistent or biased outputs, especially when analyzing textual metadata in ID documents. Understanding how tokenization affects image interpretation could improve model robustness and fairness in PAD tasks.

\section*{Acknowledgements}
This work was partially funded by the Facephi company, R\&D area and the German Federal Ministry of Education and Research and the Hessian Ministry of Higher Education, Research, Science and the Arts within their joint support of the National Research Centre for Applied Cybersecurity ATHENE.

\bibliographystyle{IEEEtran}
\bibliography{bibliography.bib}

\begin{thebibliography}{10}
\providecommand{\url}[1]{#1}
\csname url@samestyle\endcsname
\providecommand{\newblock}{\relax}
\providecommand{\bibinfo}[2]{#2}
\providecommand{\BIBentrySTDinterwordspacing}{\spaceskip=0pt\relax}
\providecommand{\BIBentryALTinterwordstretchfactor}{4}
\providecommand{\BIBentryALTinterwordspacing}{\spaceskip=\fontdimen2\font plus
\BIBentryALTinterwordstretchfactor\fontdimen3\font minus \fontdimen4\font\relax}
\providecommand{\BIBforeignlanguage}[2]{{%
\expandafter\ifx\csname l@#1\endcsname\relax
\typeout{** WARNING: IEEEtran.bst: No hyphenation pattern has been}%
\typeout{** loaded for the language `#1'. Using the pattern for}%
\typeout{** the default language instead.}%
\else
\language=\csname l@#1\endcsname
\fi
#2}}
\providecommand{\BIBdecl}{\relax}
\BIBdecl

\bibitem{Survey-FM}
M.~Awais, M.~Naseer, S.~Khan, R.~M. Anwer, H.~Cholakkal, M.~Shah, M.-H. Yang, and F.~S. Khan, ``Foundation models defining a new era in vision: A survey and outlook,'' \emph{IEEE Trans. on Pattern Analysis and Machine Intelligence}, vol.~47, no.~4, pp. 2245--2264, 2025.

\bibitem{Survey-explainability}
\BIBentryALTinterwordspacing
R.~Kazmierczak, E.~Berthier, G.~Frehse, and G.~Franchi, ``Explainability and vision foundation models: A survey,'' \emph{Information Fusion}, vol. 122, p. 103184, 2025. [Online]. Available: \url{https://www.sciencedirect.com/science/article/pii/S156625352500257X}
\BIBentrySTDinterwordspacing

\bibitem{steiner2024paligemma2}
A.~Steiner, A.~S. Pinto, and M.~T. et~al., ``{PaliGemma-2: A Family of Versatile VLMs for Transfer},'' \emph{arXiv preprint arXiv:2412.03555}, 2024.

\bibitem{liu2023improvedllava}
H.~Liu, C.~Li, Y.~Li, and Y.~J. Lee, ``Improved baselines with visual instruction tuning,'' in \emph{2024 IEEE/CVF Conference on Computer Vision and Pattern Recognition (CVPR)}, 2024, pp. 26\,286--26\,296.

\bibitem{qwen}
J.~Bai, S.~Bai, and Y.~C. et~al., ``Qwen technical report,'' \emph{arXiv preprint arXiv:2309.16609}, 2023.

\bibitem{IJCB2024-PAD}
J.~E. Tapia, N.~Damer, C.~Busch, J.~M. Espin, J.~Barrachina, A.~S. Rocamora, K.~Ocvirk, L.~Alessio, B.~Batagelj, S.~Patwardhan, R.~Ramachandra, R.~Mudgalgundurao, K.~Raja, D.~Schulz, and C.~Aravena, ``First competition on presentation attack detection on {ID} card,'' in \emph{IEEE Intl. Joint Conf. on Biometrics (IJCB)}, 2024, pp. 1--10.

\bibitem{IJCB2025-PAD}
J.~E. Tapia, M.~Nieto, J.~M. Espin, A.~S. Rocamora, J.~Barrachina, N.~Damer, C.~Busch, M.~Ivanovska, L.~Todorov, R.~Khizbullin, L.~Lazarevich, A.~Grishin, D.~Schulz, S.~Gonzalez, A.~Mohammadi, K.~Kotwal, S.~Marcel, R.~Mudgalgundurao, K.~Raja, P.~Schuch, S.~Patwardhan, R.~Ramachandra, P.~Couto~Pereira, J.~R. Pinto, M.~Xavier, A.~Valenzuela, R.~Lara, B.~Batagelj, M.~Peterlin, P.~Peer, A.~Muhammed, D.~Nunes, and N.~Gonçalves, ``Second competition on presentation attack detection on id card,'' in \emph{2025 IEEE International Joint Conference on Biometrics (IJCB)}, 2025, pp. 1--10.

\bibitem{gonzalez2025forged}
S.~González and J.~E. Tapia, ``Forged presentation attack detection for {ID} cards on remote verification systems,'' \emph{Pattern Recognition}, vol. 162, p. 111352, 2025.

\bibitem{ChatGPT-iris}
P.~Farmanifard and A.~Ross, ``{ChatGPT} meets {Iris} biometrics,'' in \emph{IEEE Intl. Joint Conf. on Biometrics (IJCB)}, 2024, pp. 1--10.

\bibitem{ChatGPT-facepad}
A.~Komaty, H.~O. Shahreza, A.~George, and S.~Marcel, ``Exploring {ChatGPT} for face presentation attack detection in zero and few-shot in-context learning,'' in \emph{IEEE/CVF Winter Conf. on Applications of Computer Vision Workshops (WACVW)}, 2025, pp. 1602--1611.

\bibitem{ChatGPT-facebiom}
I.~Deandres-Tame, R.~Tolosana, R.~Vera-Rodriguez, A.~Morales, J.~Fierrez, and J.~Ortega-Garcia, ``How good is {ChatGPT} at face biometrics? a first look into recognition, soft biometrics, and explainability,'' \emph{IEEE Access}, vol.~12, pp. 34\,390--34\,401, 2024.

\bibitem{isoiec301073}
{ISO/IEC JTC 1/SC 37 Biometrics}, ``{ISO/IEC} 30107-3:2023. information technology --- biometric presentation attack detection --- part 3: Testing and reporting,'' International Organization for Standardization, Geneva, Switzerland, Standard, 2023.

\bibitem{ChatGPT-fmad}
H.~Zhang, R.~Ramachandra, K.~Raja, and C.~Busch, ``{ChatGPT} encounters morphing attack detection: Zero-shot mad with multi-modal large language models and general vision models,'' \emph{IEEE Transactions on Biometrics, Behavior, and Identity Science}, pp. 1--1, 2025.

\end{thebibliography}

\end{document}